\documentclass{article}
\usepackage[numbers]{natbib}
\usepackage {arxiv}
\usepackage{cite}
\usepackage{float}
\usepackage{amsmath,amssymb,amsfonts}
\usepackage{algorithmic}
\usepackage{graphicx}
\usepackage{textcomp}
\usepackage{booktabs}
\usepackage{amsmath}
\usepackage{subfigure}
\usepackage[table,xcdraw]{xcolor}
\usepackage{soul}
\usepackage{multirow}

\usepackage{soul}

 \title{\LARGE \bf Mining Electronic Health Records to Investigate Effectiveness of Ensemble Deep Clustering}

\author{Manar D. Samad,~
       Yina Hou~
        Shrabani Ghosh\\
Department of Computer Science\\
Tennessee State University\\
Nashville, TN, USA\\
\texttt{msamad@tnstate.edu} \\
}

\begin{document}

\maketitle

\begin{abstract}
In electronic health records (EHRs), clustering patients and distinguishing disease subtypes are key tasks to elucidate pathophysiology and aid clinical decision-making. However, clustering in healthcare informatics is still based on traditional methods, especially K-means, and has achieved limited success when applied to embedding representations learned by autoencoders as hybrid methods. This paper investigates the effectiveness of traditional, hybrid, and deep learning methods in heart failure patient cohorts using real EHR data from the All of Us Research Program. Traditional clustering methods perform robustly because deep learning approaches are specifically designed for image clustering, a task that differs substantially from the tabular EHR data setting. To address the shortcomings of deep clustering, we introduce an ensemble-based deep clustering approach that aggregates cluster assignments obtained from multiple embedding dimensions, rather than relying on a single fixed embedding space. When combined with traditional clustering in a novel ensemble framework, the proposed ensemble embedding for deep clustering delivers the best overall performance ranking across 14 diverse clustering methods and multiple patient cohorts. This paper underscores the importance of biological sex-specific clustering of EHR data and the advantages of combining traditional and deep clustering approaches over a single method.

\end{abstract}


\section{Introduction}

Electronic health records (EHRs) are treasure troves of medical history and information that can reveal new insights to optimize healthcare and clinical decision-making~\citep{Samad2018, hu2024detecting}. However, EHR data, including demographics, diagnoses, laboratory results, and medications, are used for patient care management, not for predictive modeling of patient risk. Therefore, substantial time, effort, and clinical expertise are required to prepare EHR data for machine learning (ML) and deep learning (DL)- based predictive modeling. The most common approach to ML/DL modeling of EHRs is to predict disease status using clinically available diagnostic labels, such as the International Classification of Diseases (ICD)~\citep{bhutto2024automatic,masud2023applying}. Supervised ML/DL methods are prone to overfitting, struggle when diagnostic labels are absent or limited despite large sample sizes, and are unable to stratify patient risk beyond binary disease labels. Unsupervised clustering offers a promising approach to learning from unlabeled data. However, modeling and evaluating clustering algorithms are not trivial tasks due to the lack of clear learning targets or diagnostic labels. Despite a decade of the deep learning revolution, clustering in practice remains strictly traditional, with k-means dominating healthcare informatics. In contrast, sophisticated DL methods have been proposed for image, text, graph, and video clustering benchmarks. Similar DL clustering methods have rarely been proposed or evaluated on structured tabular data, such as EHR data for patient risk stratification. To this end, this paper makes three contributions to address methodological gaps in the healthcare informatics literature. First, we extend the Gaussian Cluster Embedding Autoencoder Latent Space (G-CEALS)~\citep{rabbani2025deep}, one of the first and most recently proposed deep clustering methods for tabular data, to the healthcare domain. Second, real-world EHR data from heart failure patient cohorts are used to evaluate state-of-the-art deep clustering versus traditional clustering methods. Third, a novel ensemble of clustering is proposed to complement the strengths of traditional and deep clustering methods. 

The organization of the paper is as follows. Section \ref{sec:background} reviews the related work on traditional and deep clustering methods for EHRs and highlights the limitations of existing approaches for structured tabular data. Section \ref{sec:method} introduces the methodological foundations of deep clustering, describes clustering baselines in this study, and presents the evaluation metrics used for performance comparison. Section \ref{sec:data} describes the extraction, curation, and preprocessing of EHR data from the All of US Research Program, including cohort construction and feature selection. Section \ref{sec:results} presents the experimental setup, results, and visual analysis using t-SNE and UMAP. Finally, Section \ref{sec:conclu} concludes the paper by summarizing key findings and future directions. 

\section{Related work} \label{sec:background}
Supervised classification using diagnostic labels is ubiquitous in healthcare informatics, which depends on the availability of curated datasets with known disease labels. Known classification labels are used in deep learning methods to obtain an optimal feature space (aka an embedding) that enables state-of-the-art performance. Without known labels, learning a cluster-friendly embedding is an open problem in deep learning. Clustering of healthcare data often relies on traditional clustering methods due to their simplicity and competitive performance\citep{wang2020unsupervised, aljohani2024optimizing, karaccam2025patterns, hu2024detecting}. 

For example,~\citep{aljohani2024optimizing} has used several traditional clustering methods, including K-Means, DBSCAN, Hierarchical Clustering, Mean Shift, Affinity Propagation, Spectral Clustering, and Gaussian Mixture Models (GMM) for patient stratification in EHR data. Nichols et al.~\citep{nichols2022simulated} have compared the performance of latent class analysis (LCA), hierarchical cluster analysis (HCA), k-means (MCA-kmeans), and k-means (kmeans) clustering to investigate multimorbidity in simplified EHR data. Karaccam et al.~\citep{karaccam2025patterns} have used k-means clustering and validated findings with hierarchical clustering and PAM (partitioning around medoids) to identify two clinically significant phenotypic subgroups among patients with advanced heart failure in the EHR. Hu et al. have used k-means and Bayesian theorem-based clustering to identify patients at high risk of cardiovascular disease in EHRs~\citep{hu2024detecting}. A combination of deep and traditional methods has been proposed to cluster EHR data. A self-supervised autoencoder is first trained to obtain an embedding, which is then clustered using a standalone traditional clustering method (k-means, hierarchical clustering) to identify patient clusters with Type 2 diabetes~\citep{manzini2022longitudinal}. A multimodal autoencoder M-ClustEHR has been proposed~\citep{bampa2024m} to cluster EHR data with mixed data types to identify phenotypes in a sepsis cohort. They use multiple traditional clustering algorithms (k-means, ensembling, HDBSCAN, hierarchical) on an autoencoder-derived embedding of multimodal data. 

In contrast, deep clustering research contends that the data reconstruction loss of an autoencoder is not an optimal learning objective for clustering~\citep{Xie2016}. Shao et al.~\citep{shao2022application} have used the clustering objective of K-means in tandem with the autoencoder reconstruction loss in their Deep clustering with AutoEncoder embedding (DCAE) method to identify clusters of patients with chronic cough in the EHR. In terms of the cluster purity metric, DCAE outperforms Hierarchical Clustering and K-means on PCA-projected data, K-means applied to deep autoencoder (DAE) embeddings, and DCAE without the reconstruction loss. The performance of DCAE compared to K-means alone (on the raw data) is unknown, whereas K-means with PCA yields (referred to as the K-means baseline in the paper) the lowest cluster purity due to the oversimplified linear projection of data. Qiu et al. \citep{qiu2025deep} have proposed VaDeSC-EHR, a deep clustering framework specifically designed to model subgroups of patient survival risk in the EHR. In addition to data reconstruction and clustering losses, a Weibull mixture–based survival objective is optimized to learn a cluster-specific survival risk in patients with Crohn’s disease. VaDeSC-EHR is evaluated against survival-based methods and clustering baselines designed for survival analysis.

\begin{figure}[t]
\centering
\includegraphics[width=0.9\linewidth]{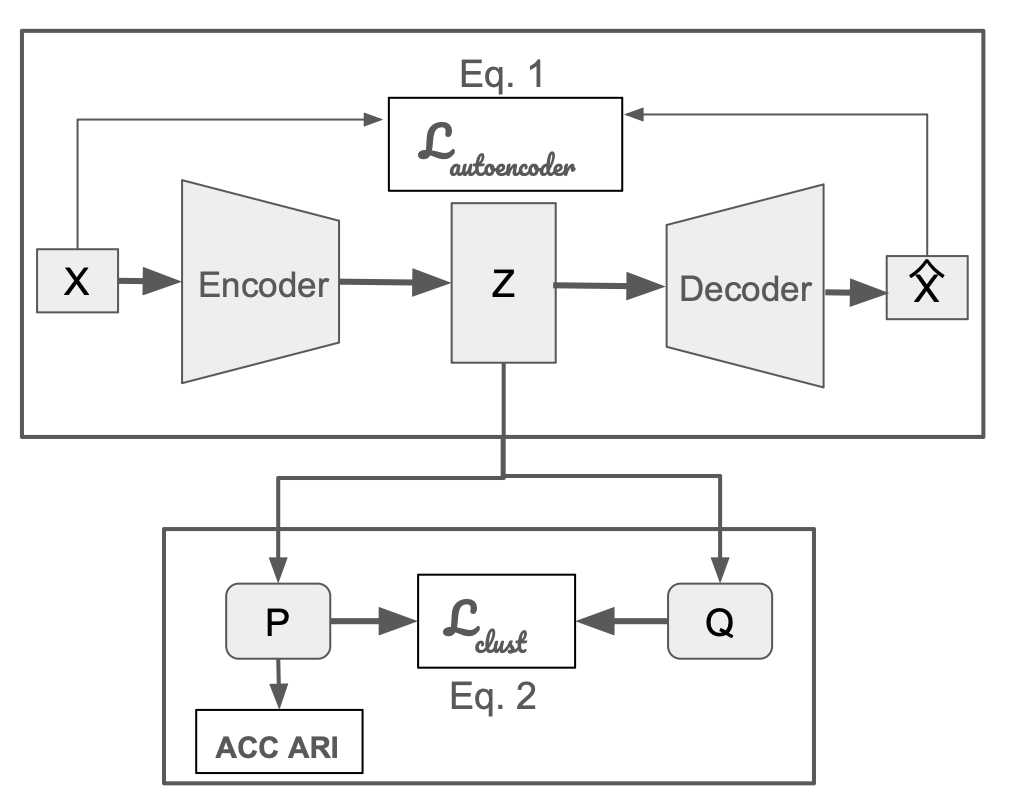}
\caption{A general framework for deep embedding clustering.}
\label{fig:gceals}
\vspace{-15pt}
\end{figure}

On the other hand, a survey on deep clustering methods~\citep{zhou2022comprehensive} does not present applications of tabular data sets similar to EHR data. Deep clustering methods are conventionally proposed for text, image, speech, video, and graph data. Deep Embedded Clustering (DEC)~\citep{Xie2016} is one of the earliest methods that pretrains an autoencoder by minimizing reconstruction loss. The encoder part is then finetuned using a clustering loss to yield a cluster-friendly embedding. The new embedding is then clustered using k-means. Later, a joint learning mechanism is proposed for the improved DEC (IDEC)~\citep{Guo2017} method. Similar to DEC and IDEC, Deep Clustering via Joint Convolutional Autoencoder (DEPICT)~\citep{Dizaji2017} is also proposed and evaluated using image data sets. DEPICT achieves embedding and clustering assignments within a single framework without requiring a standalone clustering method. Deep k-means (DKM)~\citep{MoradiFard2020} and AE-CM~\citep{Boubekki2021} use the same architecture as DEC and IDEC, but replace the embedding size from 10 to the number of clusters (k). The dynamic autoencoder (DynAE)~\citep{Mrabah_neunet_2020} uses the same learning architecture, with the objective function updated via image augmentation methods. All baseline deep clustering methods as of today are optimized for image data sets alone. A small study~\citep{abrar2023} has demonstrated that deep image clustering methods are not as effective for tabular data sets as traditional clustering methods. In contrast, a study~\citep{kowsar2023deep} on simulated EHR data (the MIMIC-III database from the Health Gym project~\citep{Kuo2021}) shows that properly tuning the weight of the IDEC clustering loss can, in certain situations, outperform k-means. To this end, G-CEALS~\citep{rabbani2025deep} is one of the first deep clustering methods proposed for tabular data, outperforming existing clustering baselines on publicly available tabular data sets. 
However, G-CEALS performance is sensitive to embedding size. Furthermore, the effectiveness of G-CEALS and other deep clustering methods in real-world EHR data remains unknown.

\section {Methodology} \label{sec:method}

This section provides a summary of the general concepts of deep clustering and the cluster evaluation methods that relate to this paper.

\subsection {Preliminaries on deep clustering}

Deep clustering methods are proposed to learn a cluster-friendly embedding that maximizes the separation between underlying data clusters. Although learning such an embedding is trivial using  known target labels, clustering of unlabeled samples remains challenging. Deep clustering frameworks start by learning a low-dimensional embedding (Z) of input data (X) using an autoencoder, as follows. 
\begin {eqnarray}
\mathcal{L}_{autoencoder} &=& \underset{\theta, \Phi}{\operatorname{argmin}} \sum^{N}_{i=1} ||  X_i - \hat{X_i} ||_2^2 \label{eq:1}
\end{eqnarray}
Here, $\theta$ and $\Phi$ are encoder and decoder parameters subject to optimization for the task of reconstructing X as $\hat{X}$ from Z. Applying traditional clustering (e.g., k-means) to embedding (Z) leverages deep learning. However, isolated embedding learning and clustering do not satisfy the definition of deep clustering. In particular, the Z space in the autoencoder is optimized to retain all information from X, particularly facilitating data reconstruction at the decoder level. Therefore, the autoencoder embedding, optimized for data reconstruction, is not theoretically optimized for cluster separation. In a deep clustering method, the pretrained autoencoder embedding is further finetuned using a carefully designed clustering loss. First, the cluster distribution of the Z space is denoted as S. Second, a target cluster distribution (T) is defined as T, which is often mathematically derived from S as a closed form solution. The general learning objective for cluster-friendly embedding is to minimize the divergence between the T and S distributions, as shown below.
\begin {eqnarray}
\mathcal{L}_{clust}  &=& \sum_{i=1}^M \sum_{j=1}^K T_{ij} log \frac{T_{ij}}{S_{ij}}\label{eq:2}
\end {eqnarray}
Here, for each of the M samples, T and S represent the corresponding likelihood of belonging to one of the K clusters. Most deep clustering methods have modeled T and S as a Student's t-distribution~\citep{Dizaji2017, Guo2017}, while G-CEALS, proposed for tabular data, models the Z space using a multivariate Gaussian distribution. The refinement of the embedding (Z) is achieved by jointly optimizing the autoencoder reconstruction loss with a clustering loss, as below.
\begin {eqnarray}
\mathcal{L}_{dc}  &=& \sum^{N}_{i=1} ||  X_i - \hat{X_i} ||_2^2  + \gamma*\sum_{i=1}^M \sum_{j=1}^K T_{ij} log \frac{T_{ij}}{S_{ij}}\label{eq:3}
\end {eqnarray}
The parameter ($\gamma$) regulates how the clustering loss ($\mathcal{L}_{cluster}$) contributes to finetuning the embedding in joint learning.

\subsection {Deep versus traditional clustering methods}

We evaluate the performance of three types of clustering approaches. First, traditional methods include direct clustering of raw data (X) using k-means and Gaussian mixture models (GMM). Second, a hybrid approach performs k-means and GMM clustering on an embedding (Z) learned by optimizing a data reconstruction loss of an autoencoder. Third, state-of-the-art deep clustering methods, including DEC~\citep{Xie2016}, IDEC~\citep{Guo2017}, DEPICT~\citep{Dizaji2017}, DynAE\citep{Mrabah_neunet_2020}, DKM\citep{MoradiFard2020} , AE-CM~\citep{Boubekki2021}, and G-CEALS~\citep{rabbani2025deep}, learn cluster assignments jointly with optimizing a cluster-friendly embedding. Notably, all deep clustering methods, except G-CEALS, are benchmarked on image data sets, and need some adjustments to enable clustering of structured tabular data of EHR. An advantage of deep clustering methods over traditional k-means is the availability of several data-specific tunable parameters, including the regularization factor and embedding size. However, optimizing these parameters for clustering methods is not standard or trivial in practice. The heterogeneity of tabular and EHR data, including varying sample sizes, mixed-type features, a wide range of distributions, and domain-specific applications, requires data-specific optimization, where one method may not be the optimal choice for all.  To this end, previous studies~\citep{rabbani2025deep} have shown that no single clustering algorithm consistently outperforms others in all types and data scenarios. 

\subsection{Cluster evaluation metrics}

Cluster assignments are compared with the true HF diagnosis labels based on ICD-10 codes to evaluate clustering performance. In particular, diagnostic labels are only used to evaluate the final cluster assignments and are never involved in training or tuning the clustering methods. However, group assignments or the choice of label 0 versus 1 to identify the group can vary between methods, which may not guarantee a one-to-one match with the true annotations of patient cohorts. Therefore, the standard classification accuracy metric would not be a valid choice for evaluation. We use three widely used cluster evaluation metrics: Accuracy (ACC), Normalized Mutual Information (NMI), and Adjusted Rand Index (ARI). Each metric captures different aspects of clustering quality with varying interpretations of cluster assignments.

\begin{itemize} 
    \item \textbf{ACC:} Clustering accuracy (ACC) uses the Hungarian algorithm~\citep{Kuhn1955} to determine the mapping m(.) between the true and predicted labels, as follows.   
\begin{equation}
ACC  = \underset{m}{\operatorname{max}} \frac{\sum_{i=1}^{N} 1 \{Y_{true}^i = = m(Y_{pred}^i)\}}{N}
 \label{equation-ACC}
\end{equation}
Here, N denotes the number of samples and m(.) finds the best mapping between the true and predicted cluster labels for each sample. Because clustering methods use 0 or 1 to label a cluster, the mapping that results in the highest alignment is used to report the cluster accuracy.
\item \textbf{Normalized Mutual Information (NMI)}: NMI measures the agreement between true labels and predicted clusters from an information-theoretic perspective and is scaled between 0 and 1. The maximum value of NMI (1) is achieved when the predicted labels perfectly match the ground truth. NMI quantifies how much information about the true clusters is captured by the predicted clustering while adjusting for differences in the number and cluster size distribution between the true and predicted groupings. Let $I(G,P)$ denote the mutual information between the ground-truth labels $G$ and the predicted cluster labels $P$. Additionally, $H(G)$ and $H(P)$ denote the entropies of the $G$ and $P$ labels, respectively, which quantify the uncertainty or randomness in the cluster labels. High entropy occurs when the samples are evenly distributed across clusters, while low entropy occurs when most samples fall into a single cluster. The NMI score is computed as follows.
\[
\text{NMI}(G, P) = 2* \frac{I(G,P)}{{H(G)\, H(P)}}
\]
As NMI is based on entropy, it is independent of the numeric permutation of class or cluster labels. Additionally, changing the true and predicted labels does not change the score \citep{estevez2009normalized}. 
\item \textbf{Adjusted Rand Index (ARI)}: ARI quantifies the similarity between the true and predicted clusters while considering chance, as shown in Eq.~\ref{eq:eval_ari}. 
\begin{equation}
\label{eq:eval_ari}
     ARI = \frac{RI - \mathbb{E}(RI)}{max(RI) - \mathbb{E}(RI)}, 
     ~~~RI = \frac{TP + TN}{{}^{N}C_2}.
\end{equation}
Here, random index (RI) uses ${}^{N}C_2$ sample pairs $(i, j)$ given N samples. A true negative (TN) pair is identified when the cluster labels for the sample pair $(i, j)$ differ at the predictive-level ($p_i \neq p_j$) as well as at the ground truth level ($g_i \neq g_j$). True positive pairs match both at the prediction level ($p_i = p_j$) and the ground truth level ($g_i = g_j$). $ \mathbb{E}(RI)$ denotes the expected value of RI. ARI ranges between -0.5 and +1, whereas negative ARI scores indicate random or discordant cluster assignments. A higher positive value is indicative of strong concordance between the true and predicted clusters. ARI is known to be a robust metric for random cluster assignments and imbalanced clusters~\citep{santos2009use}.

\end{itemize}

\begin{table}[t]
\centering
\caption{List of International Classification of Diseases, 10th Revision (ICD-10) codes corresponding to different heart failure (HF) diagnoses used in this study.}
\label{tab:ICD10_hf}
\scalebox{0.85}{
\setlength{\tabcolsep}{5pt}
\renewcommand{\arraystretch}{1.05}
\begin{tabular}{ll}
\toprule
\textbf{Diagnosis} & \textbf{ICD-10 Codes} \\
\midrule
General HF & I50 \\
Systolic (congestive) HF & I50.2–I50.23 \\
Diastolic (congestive) HF & I50.3–I50.33 \\
Combined systolic and diastolic HF & I50.4–I50.43 \\
Right HF (including due to left HF) & I50.81–I50.814 \\
Biventricular / High output / End stage HF & I50.82–I50.84 \\
Other specified HF & I50.8, I50.89 \\
Unspecified HF & I50.9 \\
\bottomrule
\end{tabular}}
\end{table}

\begin{table}[t]
\caption{Summary of the data sets stratified by gender. The combined set has an equal number of male and female patient samples.}
\label{tab:data_stat}
\centering
\scalebox{0.8}{
\begin{tabular}{llcccc}
\toprule
\textbf{Group} & \textbf{Samples} & \textbf{Study} & \textbf{Features} & \textbf{Missing} & Class Ratio\\
&&Samples&&Rate&(HF: Non-CVD)\\
\midrule
Male & 7,333 & 7333 & 33 & 0.52\% & 1 : 1.9 \\
Female & 15,346 & 7333 & 33 & 0.97\% & 1 : 1.9\\
Combined & 22,679 & 7332 & 34 & 0.63\% & 1 : 1.9\\
\bottomrule
\end{tabular}}
\end{table}

\begin{table}[t]
\centering
\caption{Summary of clinical and laboratory features extracted from the All of Us patient cohort with and without a heart failure diagnosis. The bound is used to determine outlier values for individual features.}
\label{tab:summary_features}
\begin{tabular}{lllll}
\toprule
\textbf{Feature Name} & \textbf{Abbr.} & \textbf{Unit} & \textbf{Range} & \textbf{Bound} \\
\midrule 
Age & Age & Y & 18 - 87 & 18 - 110 \\
Chloride & Cl & mmol/L & 81 - 120 & 80 - 130 \\
Sodium & Na & mmol/L & 110 - 153 & 110 - 160 \\
Calcium & Ca & mg/dL & 5.1 - 12.8 & 4 - 15 \\
Potassium & K & mmol/L & 2.1 - 10 & 2 - 10 \\
Urea nitrogen & BUN & mg/dL & 2 - 138 & 1 - 200 \\
Glucose & Glu & mg/dL & 30 - 842 & 20 - 1250 \\
Height & Height & cm & 121.92 - 213.36 & 120 - 230 \\
Creatinine & Cr & mg/dL & 0.3 - 19.9 & 0.1 - 20 \\
Weight & Weight & kg & 30.12 - 327.2 & 25 - 400 \\
Protein & TP & g/dL & 3.6 - 10.9 & 3 - 12 \\
Diastolic blood pressure & DBP & mmHg & 37 - 193 & 20 - 200 \\
Hemoglobin & Hb & g/L & 45 - 192 & 30 - 250 \\
Heart rate & HR & bpm & 20 - 159 & 20 - 250 \\
Systolic blood pressure & SBP & mmHg & 64 - 241 & 50 - 300 \\
MCV & MCV & fL & 56.9 - 125.5 & 50 - 130 \\
Alkaline phosphatase & ALP & U/L & 9.4 - 1866 & 0 - 2000 \\
Aspartate aminotransferase & AST & U/L & 3 - 1965 & 0 - 2000 \\
MCH & MCH & pg & 16.2 - 40 & 15 - 40 \\
Erythrocyte distribution width & RDW & \% & 10.7 - 33.3 & 8 - 40 \\
Carbon dioxide & CO2 & mmol/L & 10 - 45 & 10 - 45 \\
Hematocrit & HCT & \% & 16.9 - 59.7 & 10 - 70 \\
Platelets & PLT\# & $10^3$/µL & 11 - 896 & 10 - 1000 \\
Leukocytes & WBC\# & $10^3$/µL & 0.76 - 83 & 0.5 - 100 \\
Alanine aminotransferase & ALT & U/L & 3 - 1873 & 0 - 2000 \\
Erythrocytes & RBC\# & $10^6$/µL & 1.75 - 7.09 & 1 - 8 \\
Respiratory rate & RR & bpm & 6 - 48 & 4 - 60 \\
Lymphocytes percent & LYMPH\% & \% & 0 - 94 & 0 - 95 \\
Basophils percent & BASO\% & \% & 0 - 5 & 0 - 10 \\
Monocytes percent & MONO\% & \% & 0 - 27.6 & 0 - 30 \\
Triglyceride & TG & mg/dL & 20 - 1677 & 20 - 2000 \\
Eosinophils percent & EOS\% & \% & 0 - 31 & 0 - 60 \\
Cholesterol HDL & HDL-C & mg/dL & 5 - 185 & 5 - 200 \\
\bottomrule
\end{tabular}
\end{table}

\begin{table*}[t]
\centering
\caption{Clustering accuracy (ACC), Adjusted Rand Index (ARI), and Normalized Mutual Information (NMI) scores after clustering patients with and without heart failure in male, female, and combined cohorts. KGG: Ensemble of K-means, GMM, and G-CEALS (Ensemble Z) clustering using majority voting. Bold, underline, and (*) represent the best, second best, and the third best scores for a patient cohort, respectively. } 
\label{tab:results}
\vspace{5pt}
\scalebox{0.6}{ 
\centering
\begin{tabular}{llcccccccccccccc}
\toprule
Metric& Data & Kmeans & GMM & Kmeans & GMM & DKM & DEC & IDEC & AE-CM & DEPICT & DynAE & G-CEALS  & G-CEALS & G-CEALS & KGG\\
& & (X) & (X) & (Z) & (Z) &  &  &  & & & & $|Z|$ = 10 & $|Z|$ = 2 & Ensemble (Z) & Ensemble\\
 \midrule
& Male & 75.2* & 68.6 & 60.1 & 59.4 & 65.5 & 52.5 & 51.9 & 67.5 & 52.4 & 54.3 & 73.5 & 69.3 & \underline{77.2} & {\bf 79.2}\\
\multirow{3}{*}{ACC} & Female & 72.2 & \underline {76.8} & 66.2 & 74.8* & 65.4 & 67.6 & 67.7 & 66.1 & 68.0 & 60.3 & 74.4 & 70.3 & 74.8 & {\bf 78.7}\\
& Combined & 68.7* & 68.4 & 50.0 & 60.4 & 65.5 & 50.7 & 50.3 & 63.6 & 50.5 & 50.0 & 51.4 & {\bf 69.5} & 53.8 &  \underline{68.8}\\
\midrule
& Male & 0.234* & 0.089 & 0.027 & 0.032 & 0.001 & 0.002 & 0.001 & 0.045 & 0.001 & 0.003 & 0.205 & 0.089 & \underline{0.27} & {\bf 0.321}\\
\multirow{ 3}{*}{ARI} & Female & 0.159 & \underline{0.232} & 0.056 & 0.108 & 0.000 & 0.122 & 0.123 & 0.038 & 0.126 & 0.023 & 0.220 & 0.123 & 0.223* & {\bf 0.315}\\
& Combined & {\bf 0.129} & 0.086 & -0.007 & 0.033 & 0.001 & 0.000 & 0.000 & 0.001 & 0.000 & 0.000 & 0.000 & 0.102* & 0.004 & \underline{0.128}\\
 \midrule
& Male & 0.144* & 0.046 & 0.007 & 0.014 & 0.001 & 0.001 & 0.000 & 0.037 & 0.000 & 0.000 & 0.119 & 0.061 & \underline{0.192} & {\bf 0.225}\\
\multirow{3}{*}{NMI} & Female & 0.071 & 0.111 & 0.013 & 0.032 & 0.000 & 0.081 & 0.078 & 0.015 & 0.078 & 0.005 & 0.132* & 0.070 & \underline{0.138} & {\bf 0.211}\\
&Combined & {\bf 0.068} & 0.043 & 0.007 & 0.010 & 0.001 & 0.000 & 0.000 & 0.000 & 0.000 & 0.000 & 0.002 & 0.058* &	0.0011 & \underline{0.065}\\
\midrule 
Avg. Rank& &3.0(2.0) & 3.8(1.8) & 11.3(2.5) & 7.7(2.5) & 11.5(3.5) & 10.3(1.2) & 10.5(2.8) & 8.8(2.8) & 10.0(3.6) & 11.3(1.5) & 6.3(4.0) & 5.3(2.3) & 3.7(2.1) & 1.3(0.6) \\
\bottomrule
\end{tabular}}
\vspace{-5pt}
\end{table*}

\begin{table*}[t]
\centering
\caption{Computational time in seconds required to complete clustering of the male patient cohort across all baseline and proposed methods. Kmeans (Z) and GMM (Z) include autoencoder pretraining time. G-CEALS Ensemble (Z) reflects the average time across all embedding dimensions from $d_k$ = 2:3:33.} 
\label{tab:time}
\vspace{5pt}
\scalebox{0.75}{ 
\centering
\begin{tabular}{ccccccccccccc}
\toprule
Kmeans & GMM & Kmeans & GMM & DKM & DEC & IDEC & AE-CM & DEPICT & DynAE & G-CEALS  & G-CEALS & G-CEALS \\
 (X) & (X) & (Z) & (Z) &  &  &  & & & & $|Z|$ = 10 & $|Z|$ = 2 & Ensemble (Z)\\
 \midrule
  1.04 & 0.68 & 300.65 & 304.17 & 1133.4 & 62.88 & 179.26 & 561.3 & 81.02 & 12476.94 & 2373.33 & 2356.4 & 2190.27 \\
 \bottomrule
\end{tabular}}
\vspace{-5pt}
\end{table*}

\section {Results} \label{sec:results}

All experiments using All of Us research data must be conducted within the program’s secure cloud-based Researcher Workbench to protect the privacy of All of Us participants. We provision a cloud computing environment with a 2-core CPU with 13 GB of RAM. Traditional clustering methods are implemented using the scikit-learn package. For deep clustering methods, DynAE, DKM, and AE-CM are developed using TensorFlow. DKM, DEC, IDEC, and DEPICT are developed using the PyTorch package.

\subsection{EHR data extraction and curation} \label{sec:data}

For this study, male and female cohorts with and without heart failure (HF) diagnosis are extracted and curated from the All of Us Research Program sponsored by the US National Institutes of Health (NIH)~\citep{allofus}. The All of Us Research Program is a nationwide cohort of more than one million volunteers who contributed biospecimens, physical measurements, and extensive health and lifestyle surveys~\citep{sankar2017precision} to facilitate health science research. The cohort of patients with an HF diagnosis is identified using multiple ICD-10 codes in I50, as listed in Table~\ref{tab:ICD10_hf}. The search results in 578,309 medical records for 17,994 unique patients with one or more forms of HF diagnosis. The contrasting cohort of patients without cardiovascular disease (non-CVD) is identified using ICD-10 codes that exclude I00-I99, which cover diseases of the circulatory system~\citep{griffiths2004impact}. The search identifies 18,554,437 medical records of 175,530 unique patients. Following the extraction of cohorts of patients, the clinical variables associated with HF are extracted, using several relevant studies~\citep{Luo2022, Carr2020, Zhu2023, Huang2023}. The 33 selected variables are listed in Table~\ref{tab:summary_features}, based on their clinical importance and frequency in routine clinical care. In particular, a patient usually has multiple follow-up records in the EHR. For individual patients, we select the first chronological medical record with a diagnosis of HF and exclude subsequent medical records. Real-world EHR data are always riddled with outliers and missing value problems. We exclude extreme outliers that exceed plausible values observed in clinical practice following the bounds presented in Table~\ref{tab:summary_features}. Furthermore, patient samples with missing values greater than 5\% are excluded to ensure minimal missing rates in the data necessary for predictive modeling. Our curated EHR data finally include a total of 22,679 patients with a missing rate of 0.63\%. Of the final patient samples, 16,478 represent the non-CVD cohort and 6201 with one or more HF diagnoses. The total missing rate for the curated data is 0.63\%. Missing values are imputed using the median value of each variable. We factor in sex as a biological variable because the prognosis of HF varies differently between men and women~\citep{tsao2023heart}. Of 22,679 patients, the male cohort includes 7,333 patients with 4,799 non-CVD and 2,534 HF at a missing rate of 0.52\%. For a fair comparison between cohorts, we sample an equal number (7,333) of patients from the remaining 15,346 female patients. The non-CVD and HF ratio in the female cohort is kept the same as in the male cohort (non-CVD 4,799 and HF 2,534) to eliminate the effect of varying clustering imbalances. A combined cohort is obtained by sampling the same number of patients (7,333) from 22,679, with equal proportions of males and females, and a missing rate of 0.63\%. The non-CVD-to-HF sample ratio in the combined cohort is the same as in the individual patient cohorts. Table~\ref{tab:data_stat} presents a summary of the patient cohort used in this study.
\subsection{Clustering model setup}
We implemented the traditional clustering baselines, K-means and GMM, using the default parameter settings in Scikit-learn. For deep learning approaches, we have employed distinct architectures and training schedules. The DEC, IDEC, and DynAE methods utilize a fully connected autoencoder with the architecture $h-500-500-2000-d-2000-500-500-h$, where $h$ represents the input dimension and $d$ is the embedding dimension, which is set to 10 for the above models. The DKM and AE-CM methods use the same structure, but the embedding size is set to the number of target clusters. These deep clustering methods are trained using the Adam optimizer with a learning rate of 0.001 and a batch size of 256. The training procedure involved pretraining the autoencoder for 1,000 epochs, followed by joint fine-tuning with the clustering loss for an additional 1,000 epochs. The default setting for G-CEALS includes an embedding dimension of 10, training for 1e4 epochs, a learning rate of 1e-5, and a batch size of 256. Table~\ref{tab:time} presents the computational time required to complete clustering of the male patient cohort across all methods.

\subsection{Model-specific performance} Table~\ref{tab:results} presents three clustering scores for different patient cohorts. Traditional K-means clustering of raw patient data (X) appears to be one of the best clustering methods with promising scores. Clustering of X by K-means yields the best ARI (0.129), NMI (0.068) scores, and the third best ACC score (68.7) for the combined cohort of male and female patients. For the male cohort, K-means achieves the third best ACC, ARI, and NMI scores with the clustering accuracy (ACC) being the second best. However, traditional clustering (K-means and GMM) in the Z embedding space shows a substantial drop in performance scores with the exception of clustering accuracy (ACC) for the cohort of female patients (ranked third among 14 clustering methods). Therefore, embedding of the autoencoder with the objective of optimizing data reconstruction loss can adversely affect cluster separation by potentially collapsing clusters in the new embedding~\citep{Xie2016}. None of the state-of-the-art deep clustering baselines (DKM, DEC, IDEC, AE-CM, DEPICT) are among the top three best performing methods for any of the patient cohorts or performance metrics. The under-performance of deep clustering methods can be attributed to their targeted design and application for benchmarking on image data sets, which are fundamentally distinct from tabular data with permutation-invariant and heterogeneous features. The deep clustering method proposed for tabular data (G-CEALS) with its default embedding dimension of 10 shows competitive clustering performance, which is superior to traditional clustering of an embedding (Z) and baseline deep clustering methods. For the female patient cohort, in particular, G-CEALS with default embedding size (10) outperforms K-means of X in all three performance metrics (ACC, ARI, and NMI). Interestingly, GMM clustering of X yields the second-best ACC and ARI scores for clustering the female patient cohort. The leading performance of GMM and G-CEALS in this context suggests that the female patient cohort data are best modeled by Gaussian clusters, which is a common attribute of both clustering methods. In contrast, when the embedding dimension is reduced from 10 to 2, the G-CEALS clustering performance decreases for both male and female cohorts of patients. However, in the embedding dimension of 2, G-CEALS yields the best ACC score for the combined patient cohorts, and the third-best ARI and NMI scores for the same patient cohort. Despite the promising performance of G-CEALS across a wide range of tabular data sets, its performance depends on the data types and is sensitive to the choice of embedding dimension. Therefore, data-specific optimization of the embedding dimensions is a key factor in the success of the G-CEALS method compared to traditional clustering methods.

\begin{figure}[t]
    \centering
     \subfigure[G-CEALS (Male)]{\includegraphics[width=0.4\textwidth]{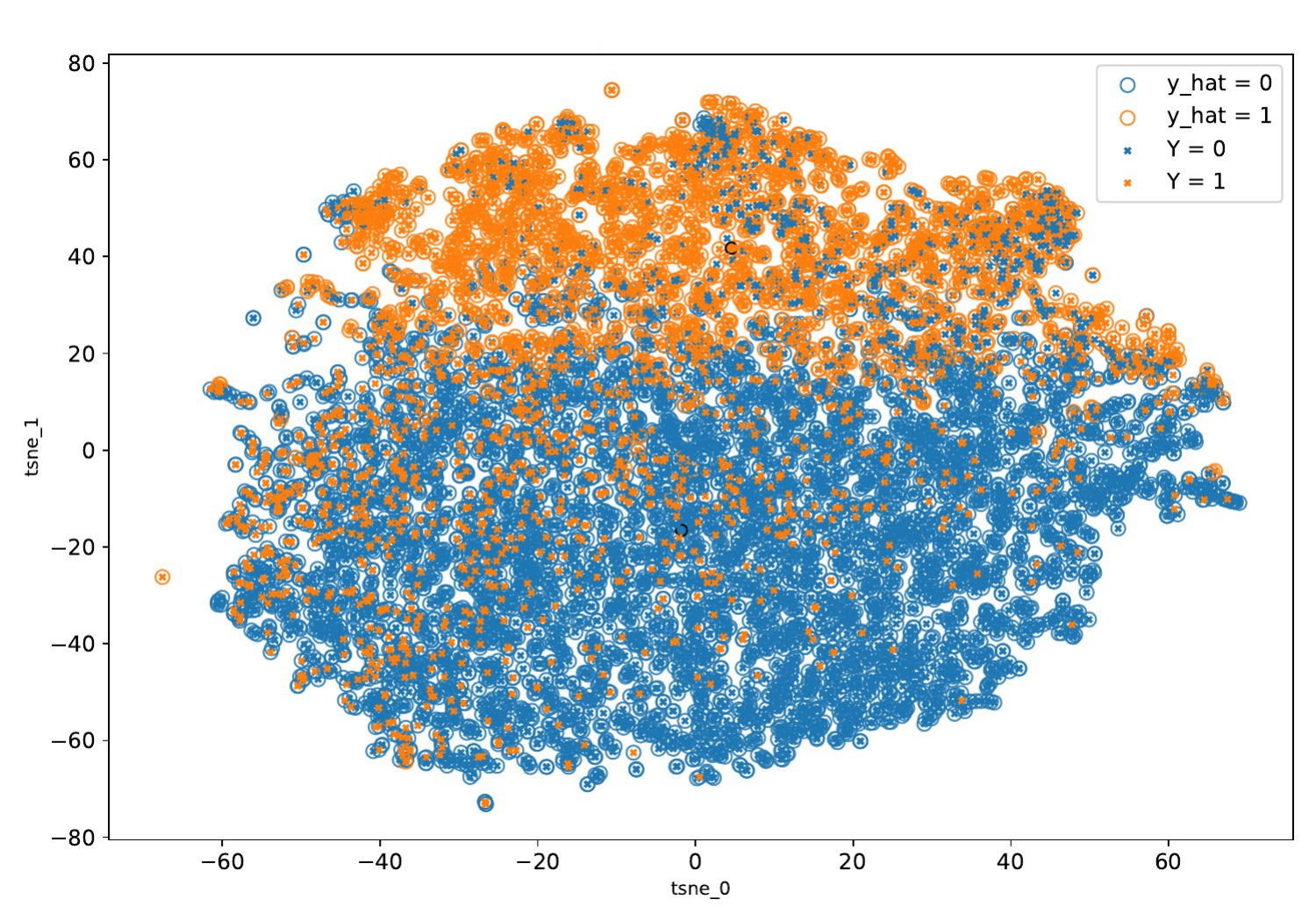}}
         \subfigure[K-means (Male)]{\includegraphics[width=0.4\textwidth]{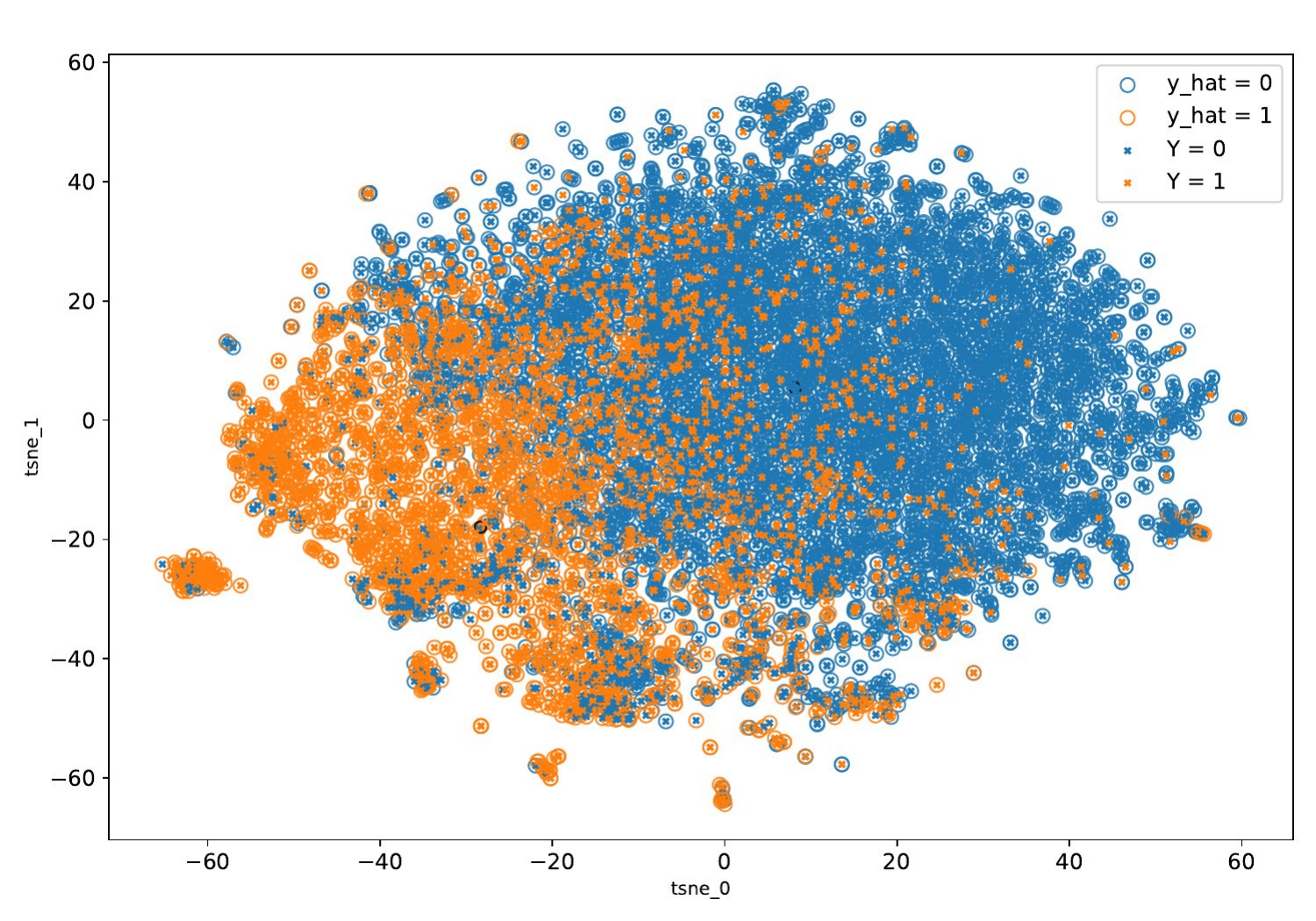}}
         \\
     \subfigure[G-CEALS (Female)]{\includegraphics[width=0.4\textwidth]{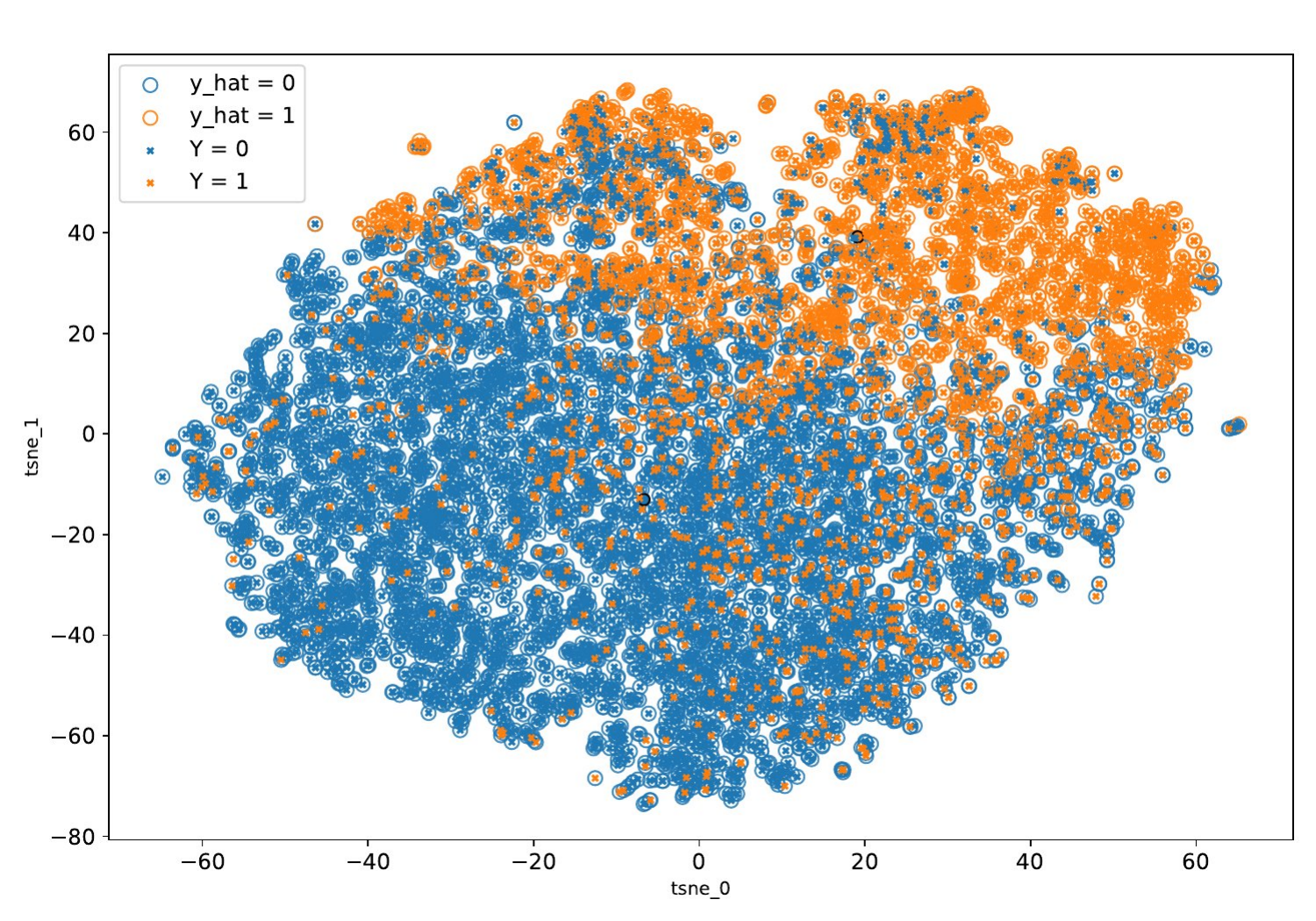}}
          \subfigure[K-means (Female)]{\includegraphics[width=0.4\textwidth]{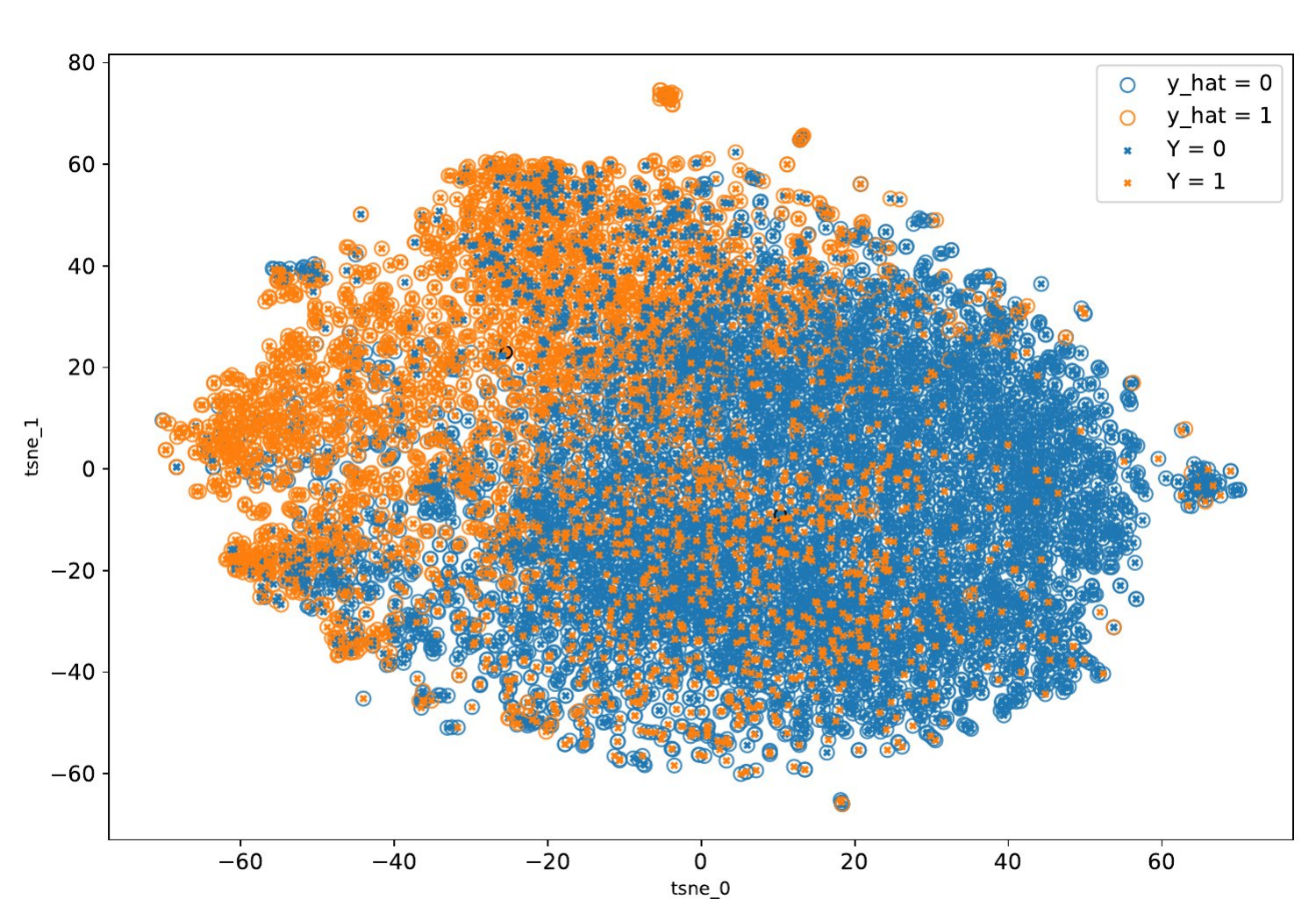}}
    \\
     \subfigure[G-CEALS (Combined)]{\includegraphics[width=0.4\textwidth]{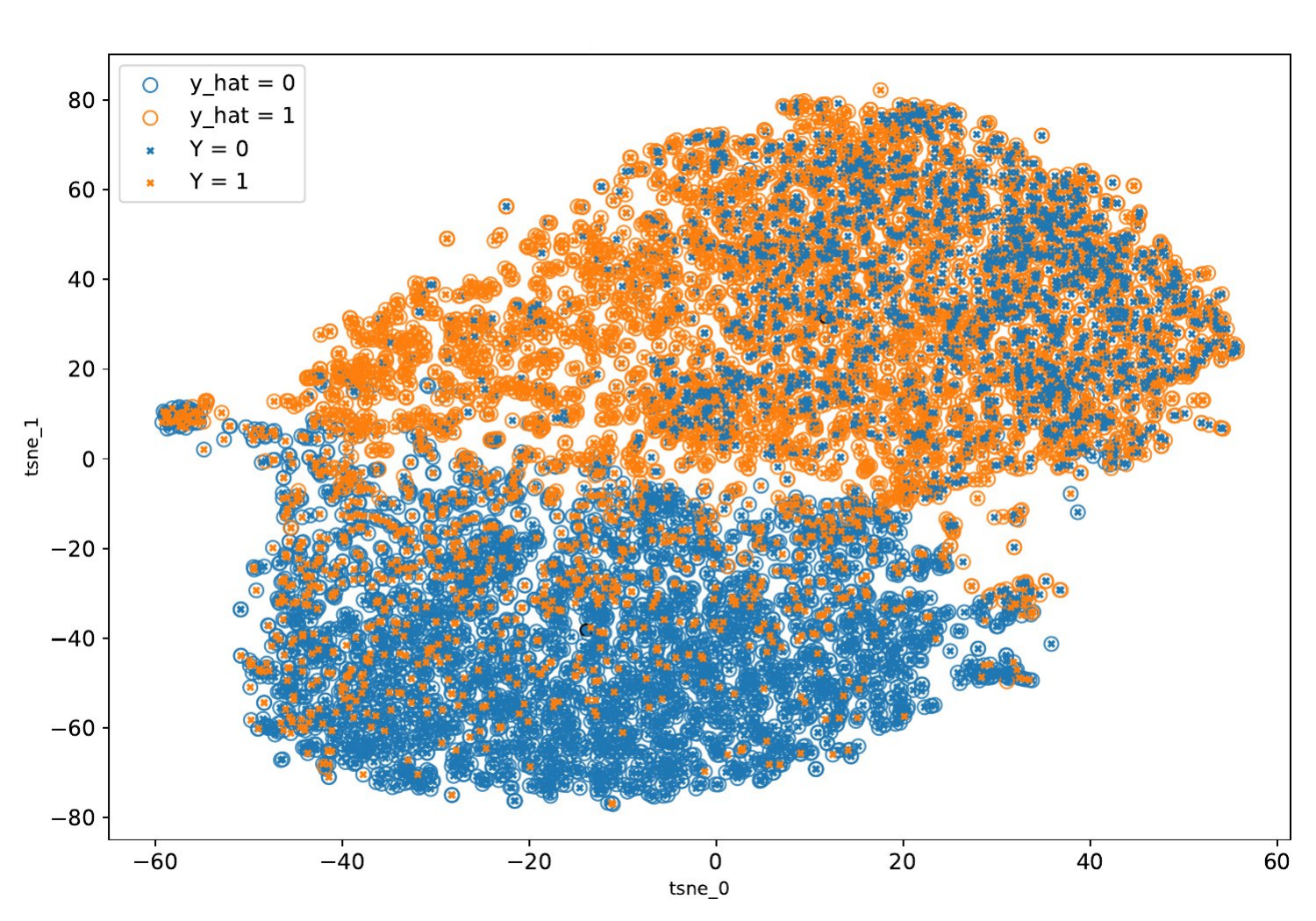}}
     \subfigure[K-means (Combined)]{\includegraphics[width=0.4\textwidth]{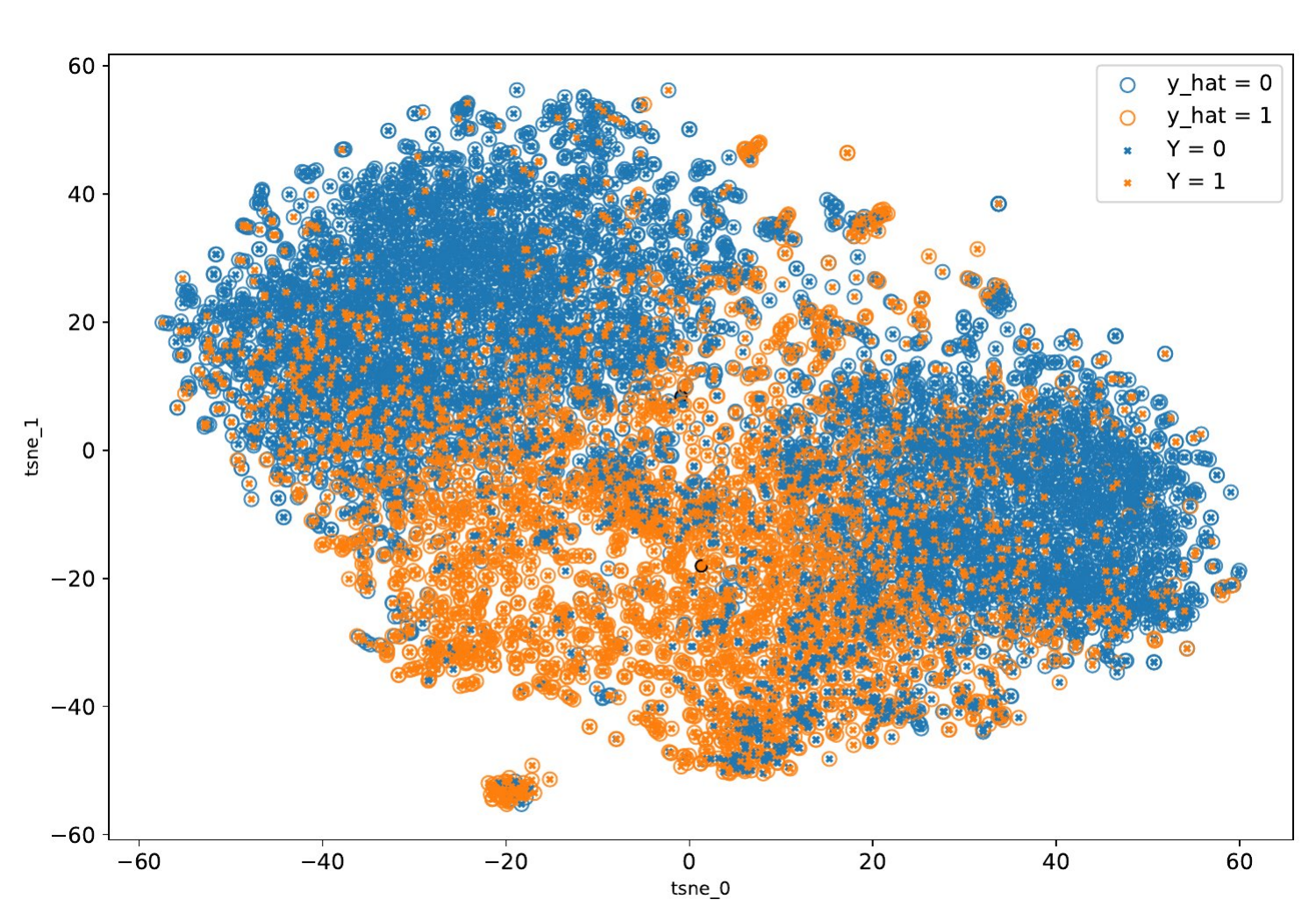}}
    \caption{Patient cluster labels color-coded on t-SNE visualization embeddings. t-SNE is applied to G-CEALS with an embedding size of 10 and to raw data to visualize K-means clusters. }
    \label{fig:tsne}
\end{figure}

\begin{figure}[t]
    \centering
    \subfigure[G-CEALS (Male)]{
\includegraphics[width=0.4\textwidth]{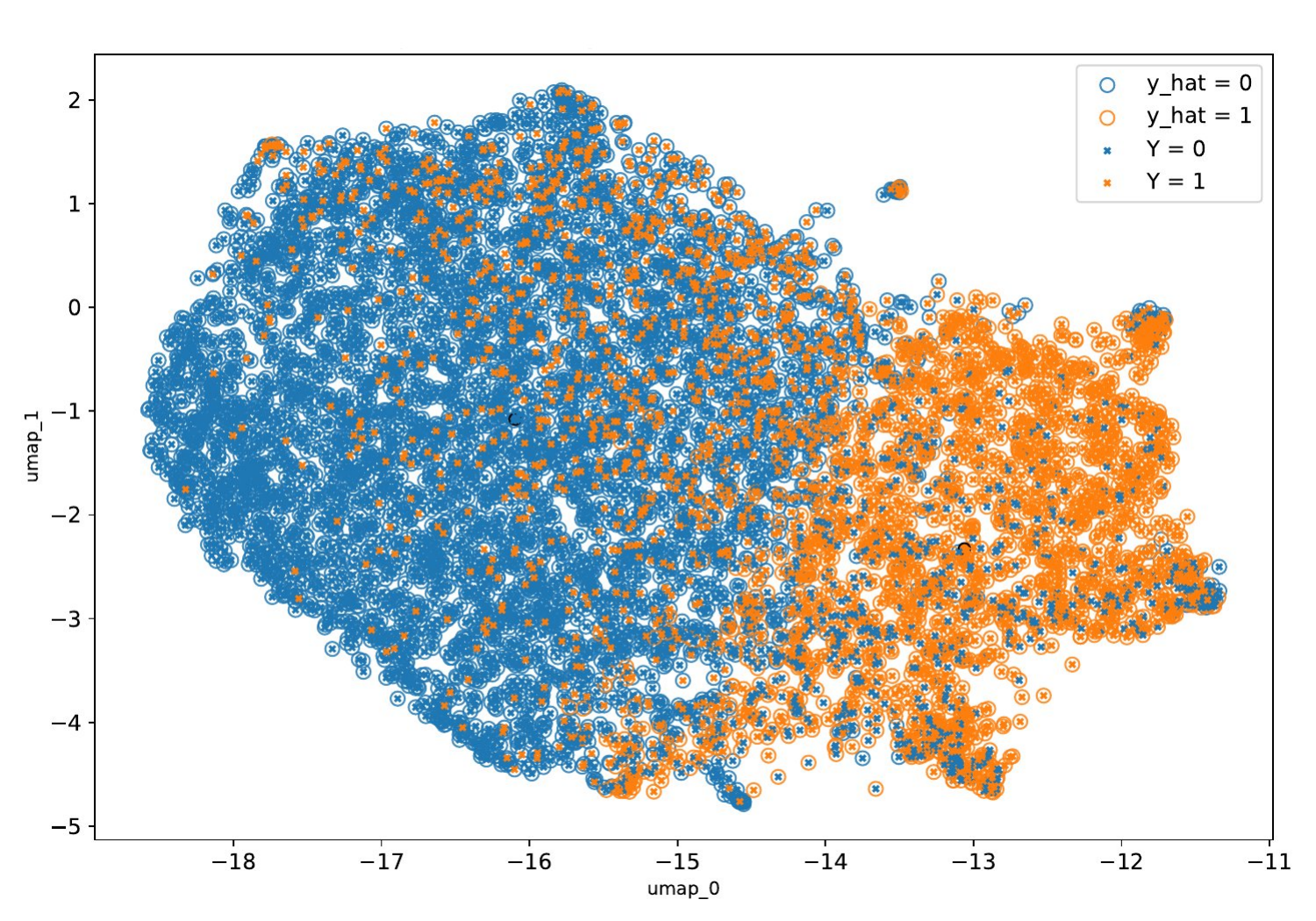}}
\subfigure[K-means (Male)]{
    \includegraphics[width=0.4\textwidth]{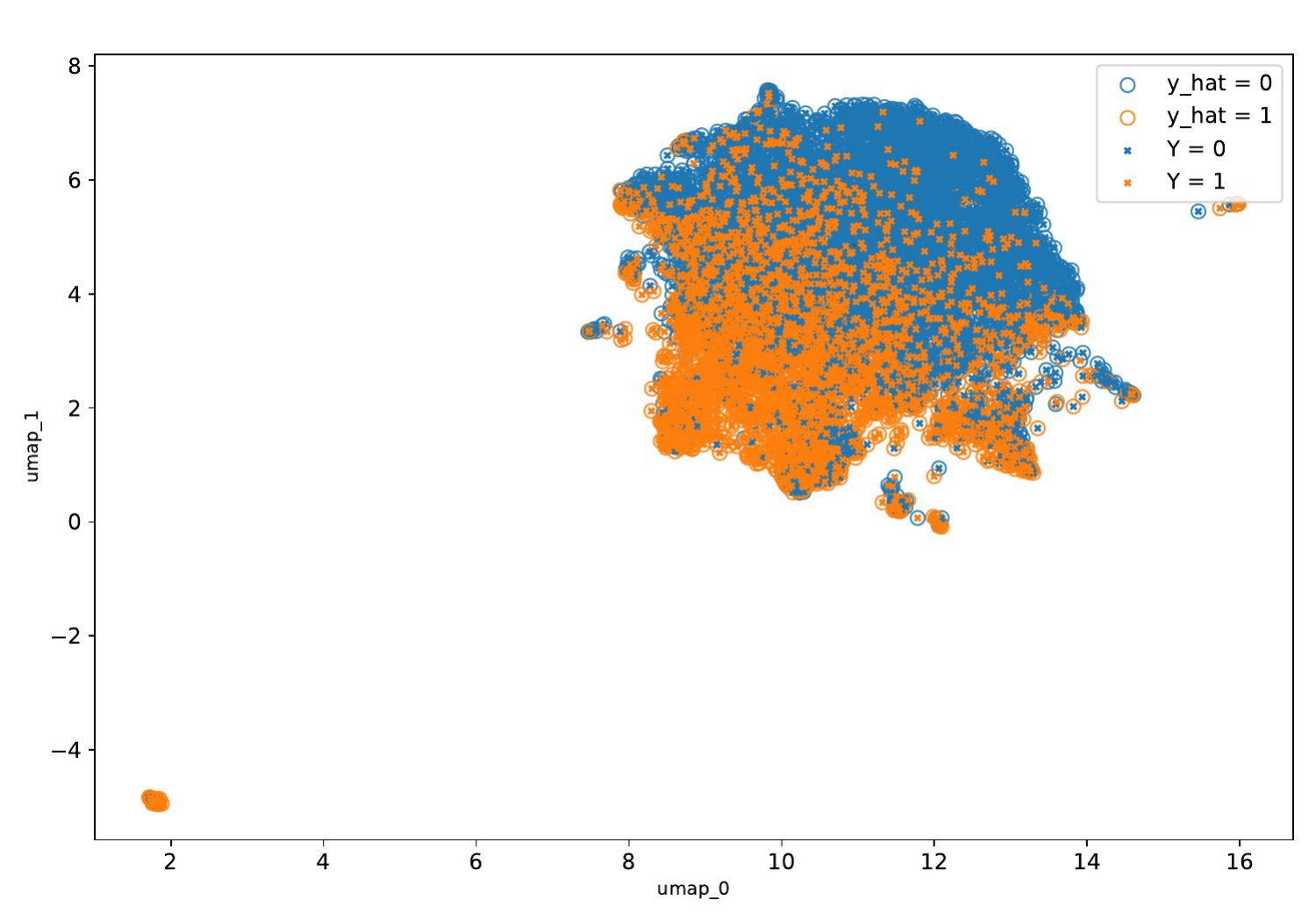}} \\
    \subfigure[G-CEALS (Female)]{
    \includegraphics[width=0.4\textwidth]{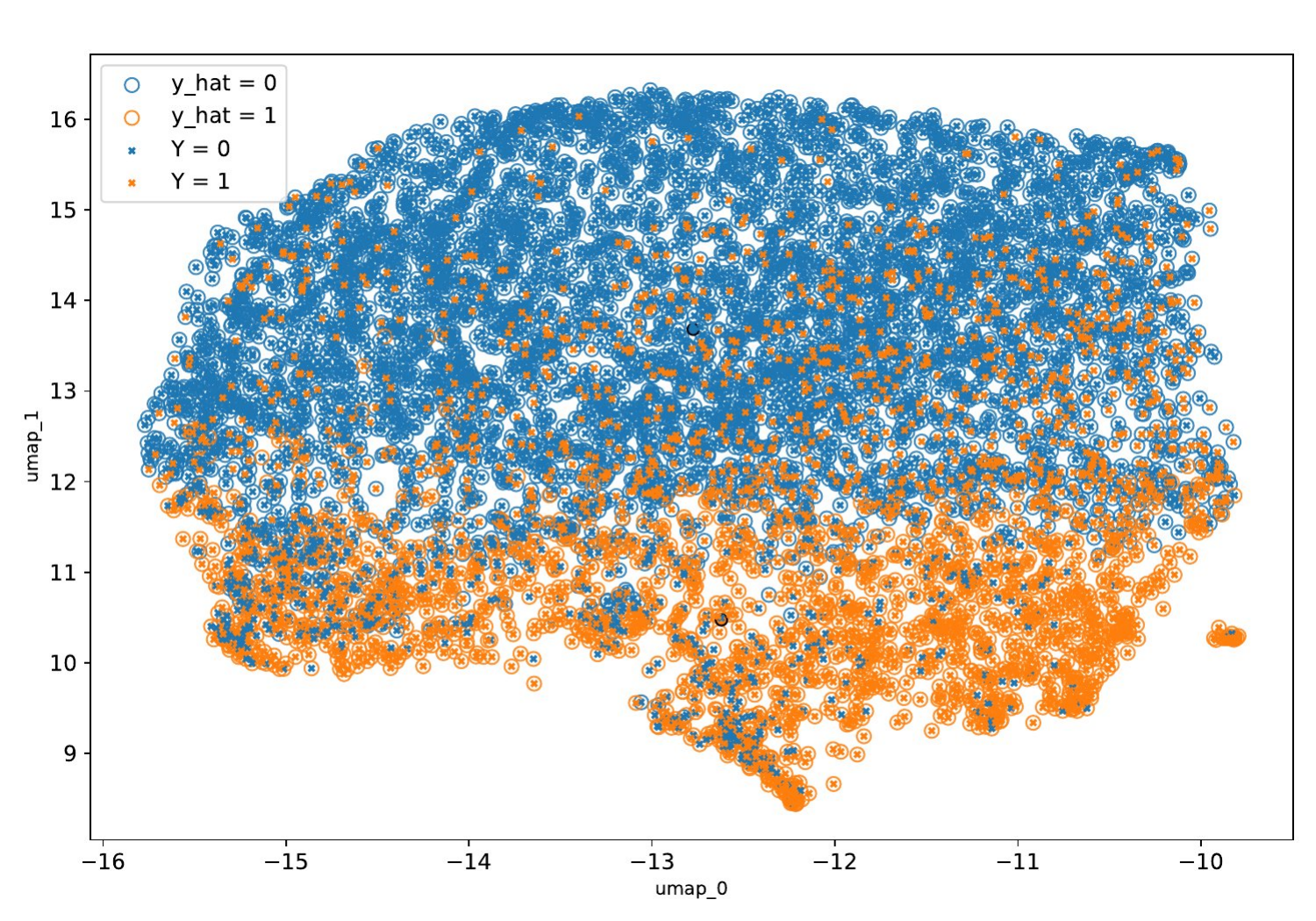}}
       \subfigure[K-means (Female)]{
    \includegraphics[width=0.4\textwidth]{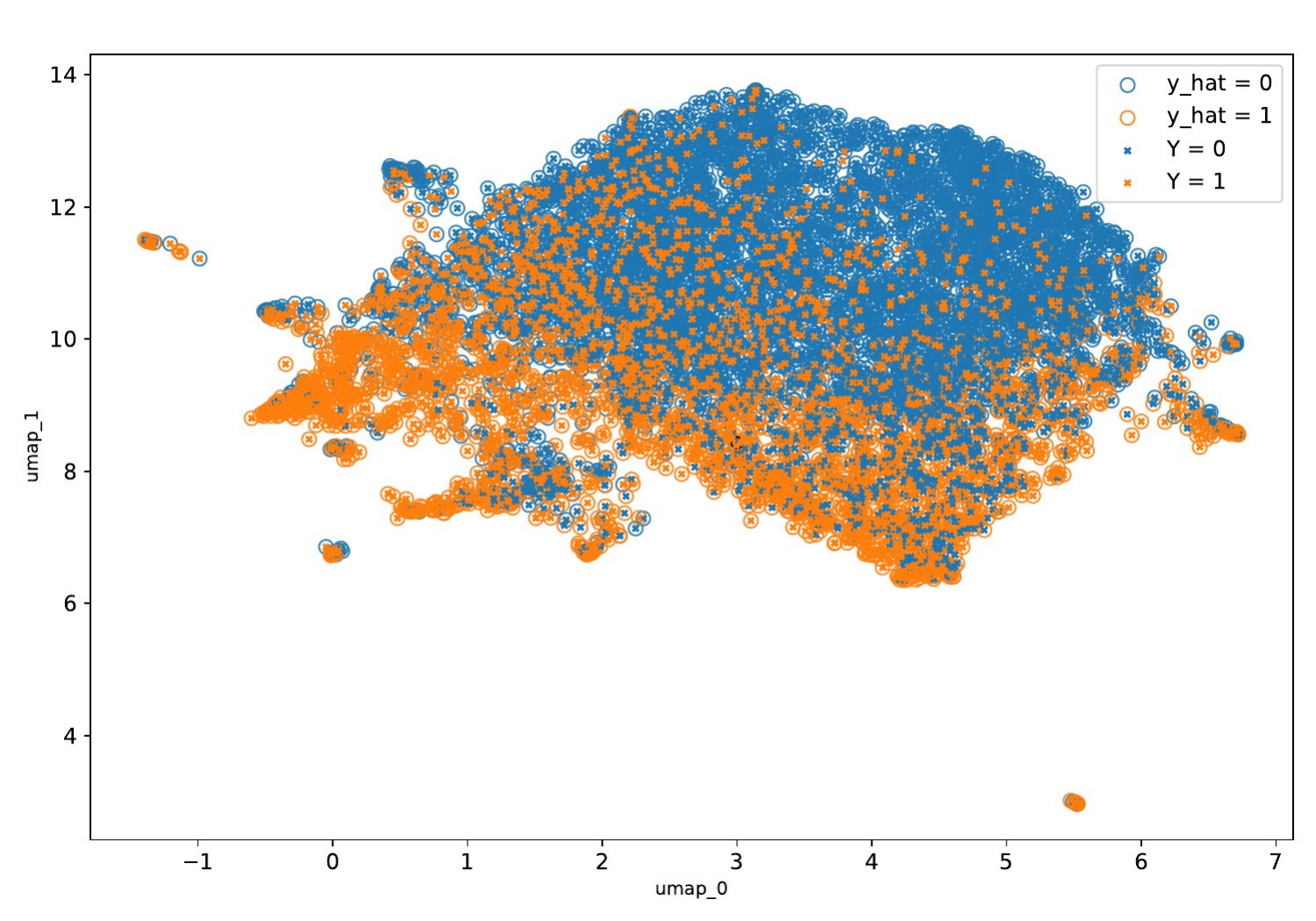}}\\
    \subfigure[G-CEALS (Combined)]{
    \includegraphics[width=0.4\textwidth]{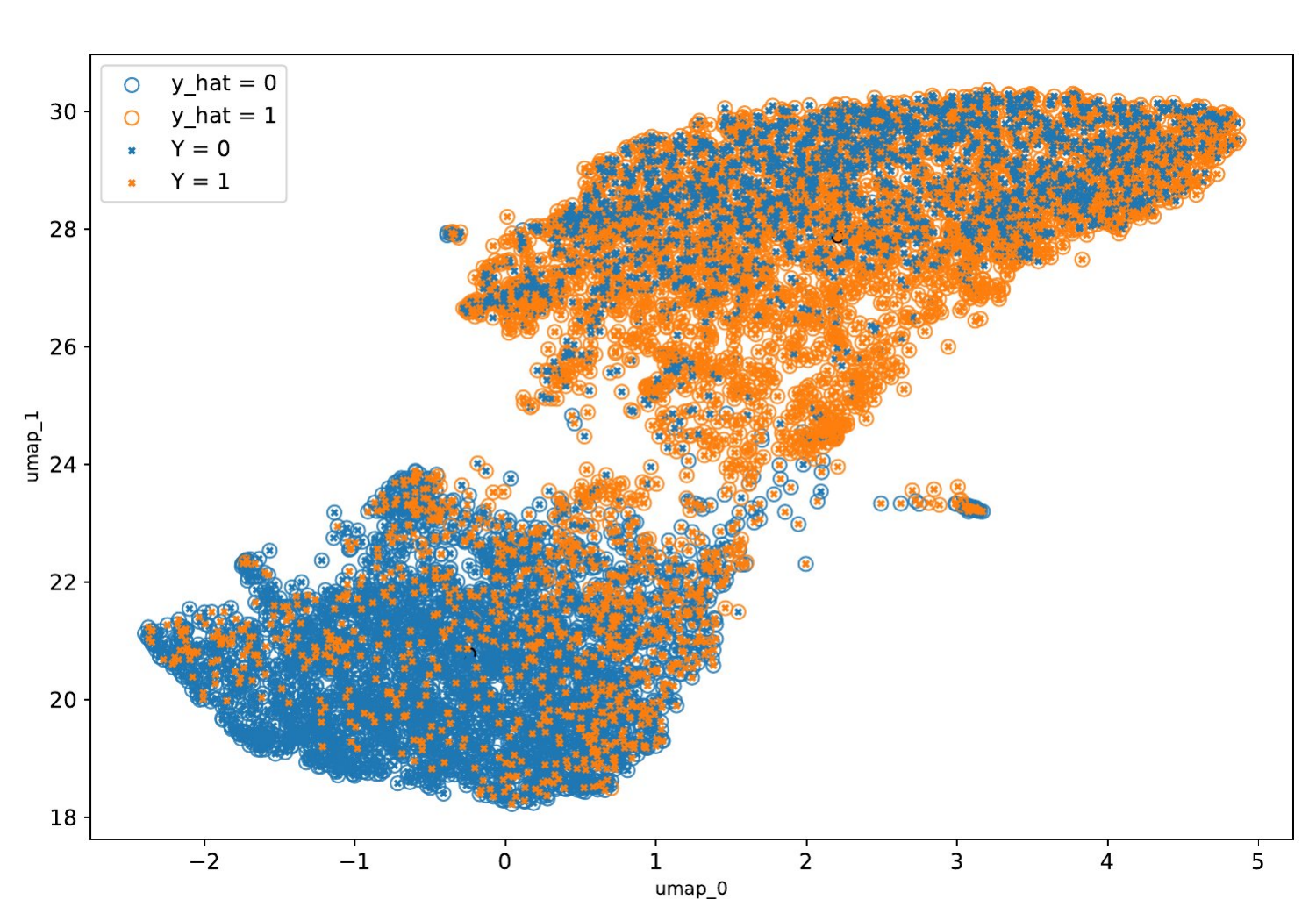}}
    \subfigure[K-means (Combined)]{
    \includegraphics[width=0.4\textwidth]{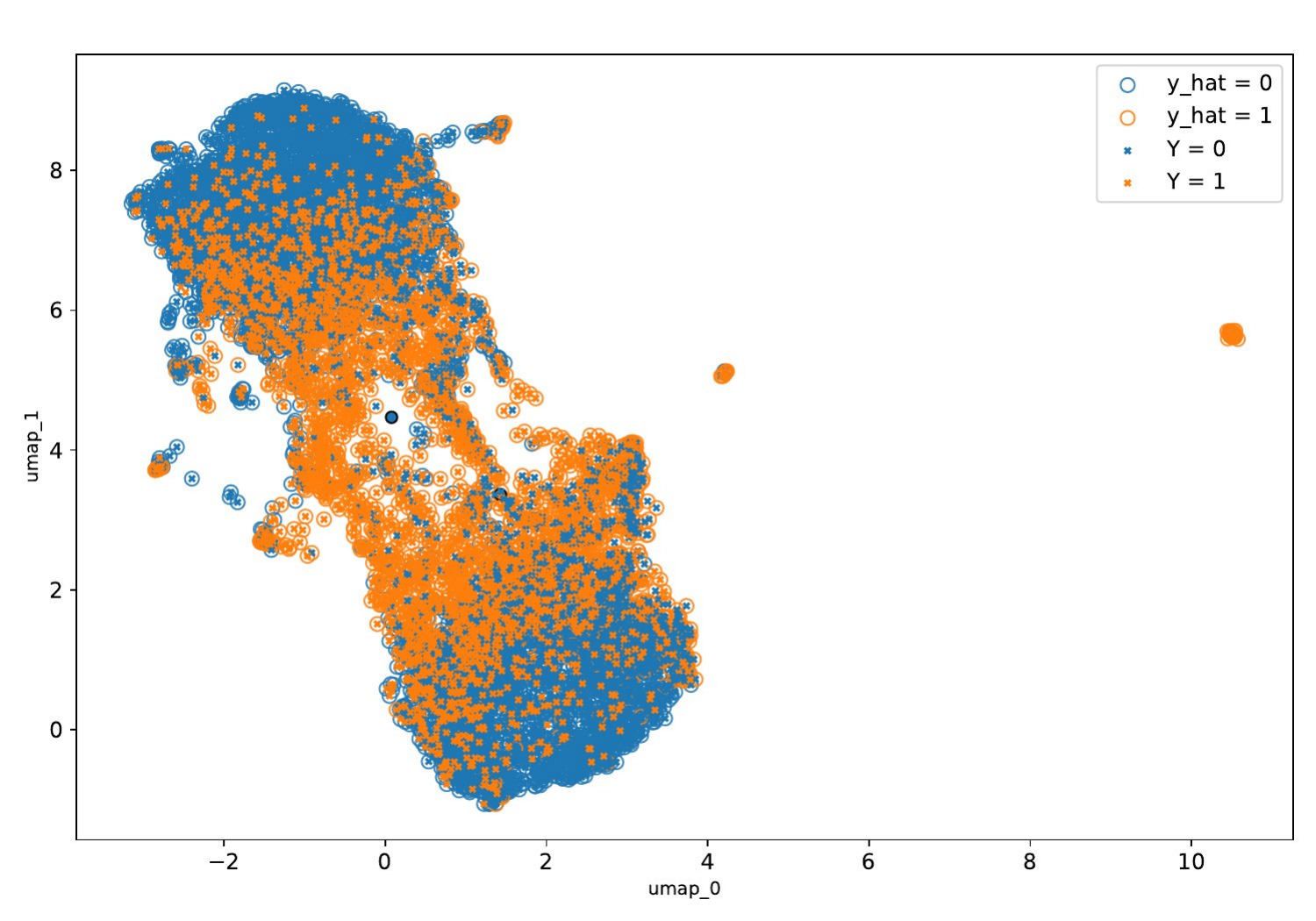}}
    \caption{Comparison of UMAP visualizations for G-CEALS (latent dimension = 10) and K-means on raw data across male, female, and combined data sets.}
    \label{fig:umap}
\end{figure}

\subsection{Ensemble of clustering}

As shown in the previous section, the embedding dimension of G-CEALS affects clustering performance. This implies that different patient subsets would be best clustered in different embedding sizes. Consequently, we spaced the embedding sizes from 2 to the full dimension of the raw feature space (X) with a step size of three, as $D$ = 2:3:33 = [$d_k$]. We obtain the cluster labels (\emph{$L_{d_k}$} = [\emph{$l_{k, j}$}]) for a distinct embedding size ($d_k$). When the concordance between the reference labels ($L_{ref}$) and the labels generated by model k is at least 50\% of the total samples, we retain the labels. Otherwise, the binary labels are flipped to harmonize most labels with the reference labels.
\begin{equation}
\hat{l}(k,j) =
\begin{cases}
l(k,j), & \text{if } \sum_{j} \mathbf{1}\{l(k,j) = l_{ref}(j)\} \geq n/2 \\[1mm]
1 - l(k,j), & \text{otherwise}
\end{cases}
\label{eq6_flip}
\end{equation}

The clustering ensemble is then obtained by averaging all cluster labels in varying embedding dimensions, as follows.
\begin{eqnarray}
    L_{avg} (j) &=& \frac{1}{|D|} \sum_{k=1}^{|D|} \hat{l} (k, j) \\
    \label{eq:lavg}
    L_{ens} (j) &=& \begin{cases}
  1, ~~ \mbox{if}~~L_{avg} (j) \geq 0.5\\    
  0, ~~\mbox {otherwise}  
  \label{eq:lens}
\end{cases}
\end{eqnarray}
Here, $j$ = $1,...,n$ and $n$ = number of samples. For clustering ensembles, we use averaging because it outperforms majority voting. Moreover, training $|D|$ number of independent G-CEALS models to obtain cluster assignments can be time-consuming. On average, the G-CEALS model with default settings takes 36 minutes to train and produce cluster assignments. The ensemble of clusters $L_{ens}$ in Equation \ref{eq:lens} from varying embedding sizes of G-CEALS is subject to the ACC, ARI, and NMI scoring. Table~\ref{tab:results} shows that the G-CEALS ensemble (Z) improves the ACC, ARI, and NMI scores for three patient cohorts compared to its default setup with $|Z| =10$. 

Across male and female cohorts, ensemble clustering of G-CEALS yields the second- or third-best ACC, ARI, and NMI scores, outperforming k-means on five of nine data and score scenarios. At this stage of the results, it is evident that traditional clustering methods remain competitive across several data scenarios. To take advantage of traditional clustering, we further aggregate the cluster labels of K-means ($L_{k-means}$), GMM ($L_{GMM}$), and G-CEALS ensemble ($L_{ens}$) (Equation \ref{eq:lens}) by majority voting, denoted as KGG ensemble ($L_{KGG}$). Prior to majority voting, the K‑means, GMM, and G‑CEALS ensemble labels are aligned to a common reference using the same flipping rule in the Equation \ref{eq6_flip}, so that all three methods vote on a consistent group identity.

Table~\ref{tab:results} shows that the KGG ensemble approach yields a substantial improvement in ACC, NMI, and ARI scores, compared to the G-CEALS ensemble and K-means for the sex-specific cohorts. The KGG ensemble achieves the highest ACC, ARI, and NMI scores for both male and female cohorts among 14 clustering methods. In the combined patient cohort, the KGG ensemble ranks second for ARI and NMI, respectively. The performance drop can be attributed to two factors. First, sex can substantially alter the distribution and separation of disease clusters, which complicates our target HF cluster separation in the learned embeddings. Second, deep clustering methods (except G-CEALS, $\|Z\|$ = 2) perform poorly in the combined cohort, where the KGG Ensemble includes the G-CEALS Ensemble instead of G-CEALS, $\|Z\|$ = 2. Since the KGG ensemble employs equal-weight majority voting across K-means, GMM, and G-CEALS Ensemble, the large inconsistency in K-means and G-CEALS performances may attenuate the diversity benefit that majority voting is designed to exploit. However, based on average performance rank across all patient cohorts and performance metrics, the KGG ensemble ranks the best, followed by K-means (X), the G-CEALS ensemble, and GMM (X).

\subsection {Visualization of cluster embeddings}
Cluster labels and separations can be explained by visualizing them in a two-dimensional (2D) space. We use t-SNE~\citep{maaten2008visualizing} and UMAP~\citep{mcinnes2018umap} to visualize cluster separation projected in 2D. Figure~\ref{fig:tsne} shows the 2D t-SNE embedding of a 10-dimensional embedding learned by G-CEALS (panels a,c,e) and the raw feature space with K-means clustering labels (panels b,d,f). Each point represents an individual patient, color-coded by ground-truth labels (Y=0 and Y=1) and predicted labels (Y\_hat=0 and Y\_hat=1). In the male cohort (ab), G-CEALS yields more diffuse samples with better separation between patient groups than the raw feature space obtained by K-means. In the female cohort (c-d), G-CEALS reveals reduced overlaps between groups of patients, while K-means labels on the raw space show substantial overlaps. For the combined (e-f), G-CEALS isolates two dominant cluster patient groups. In contrast, the separation is less pronounced when K-means clusters are applied to the t-SNE map of the raw data. In general, t-SNE plots indicate that G-CEALS yields improved visual separability and a cluster-friendly representation. Similarly, in the UMAP visualization (Figure~\ref{fig:umap}), the male cohort (a-b) and the female cohort (c-d) exhibit patterns similar to their t-SNE counterparts. However, the UMAP embedding of the raw data appears to be more concentrated than its t-SNE counterpart.

\subsection{Summary of the results and future work}

This study evaluates traditional clustering, hybrid clustering with deep embedding and deep clustering, and cluster ensemble methods to stratify patients with and without heart failure diagnoses using real-world EHR data. The key findings of the paper can be summarized as follows. First, traditional clustering methods, directly applied to raw data, remain competitive for patient stratification in EHR data. Second, embedding learned by an autoencoder, by optimizing a data reconstruction loss, is not the best choice for traditional clustering methods. Third, deep clustering methods, proposed and benchmarked on image data sets, are not optimal for tabular data sets that constitute EHR. Fourth, the performance of a clustering method often depends on the data type and distributions, as there is no best method for all data problems. Therefore, clustering must be performed separately on male and female cohorts for optimal results, as lab values and distributions differ widely due to biological sex. Fifth, proposed cluster ensembles using varying dimensions of autoencoder embedding and combining the complementary strengths of traditional and deep clustering methods can achieve state-of-the-art clustering performance on EHR data. In the future, the proposed and baseline methods need to be evaluated using other cohorts of patients with large and heterogeneous patient samples. Furthermore, extending the proposed ensemble clustering framework to identify clinically relevant subphenotypes is an important direction for future work. In particular, exploring the distinct subtypes of heart failure defined by the ICD-10 codes, including systolic, diastolic, and right-sided heart failure, can provide deeper insights into disease heterogeneity and patient risk stratification beyond the binary distinction of HF vs. non-CVD explored in this work. Unfortunately, the All of Us research platform for analyzing protected EHR data provides limited cloud computing resources for processing a large volume of EHR data. More efforts will be needed to provide the necessary computing resources when using the All of Us workbench.

\section{Conclusion} \label{sec:conclu}
This paper presents a comprehensive evaluation of traditional, hybrid, and deep clustering methods for patient stratification in a real-world EHR. The results show that traditional approaches, such as K-means, remain strong baselines. However, deep learning can obtain cluster-friendly representations that improve performance in multiple patient cohorts. Our experimental findings demonstrate that aggregating clusters from varying embedding dimensions and further combining traditional and deep clustering labels yields state-of-the-art clustering performance. In general, the study highlights that the best clustering performance is achieved when patient cohorts are analyzed separately by sex and when complementary clustering approaches are combined in an ensemble method.

\section{Acknowledgments}

The research reported in this publication received support from the US National Science Foundation (NSF) award \# 2431058.

\bibliographystyle{elsarticle-num}
\bibliography{ClusterEmbed}

\end{document}